\documentclass[letterpaper]{article} 
\usepackage{aaai2026} 
\nocopyright 
\usepackage{times}  
\usepackage{helvet}  
\usepackage{courier}  
\usepackage[hyphens]{url}  
\usepackage{graphicx} 
\usepackage{natbib}  
\usepackage{caption} 
\usepackage{algorithm}
\usepackage{algpseudocode} 
\usepackage{newfloat}
\usepackage{listings}
\DeclareCaptionStyle{ruled}{labelfont=normalfont,labelsep=colon,strut=off} 
\floatstyle{ruled}
\newfloat{listing}{tb}{lst}{}
\floatname{listing}{Listing}
\usepackage[utf8]{inputenc}
\usepackage{amsmath,amssymb,amsfonts}
\usepackage{array}
\usepackage{booktabs}
\usepackage{subcaption}
\title{Structural Plasticity as Active Inference: A Biologically-Inspired Architecture for Homeostatic Control}
\author {
    Brennen Hill
}
\affiliations{
    University of Wisconsin-Madison\\
    Madison, WI, USA\\
    bahill4@wisc.edu
}
\begin{document}

\maketitle

\begin{abstract}
Traditional neural networks, while powerful, rely on biologically implausible learning mechanisms such as global backpropagation. This paper introduces the Structurally Adaptive Predictive Inference Network (SAPIN), a novel computational model inspired by the principles of active inference and the morphological plasticity observed in biological neural cultures. SAPIN operates on a 2D grid where processing units, or cells, learn by minimizing local prediction errors. The model features two primary, concurrent learning mechanisms: a local, Hebbian-like synaptic plasticity rule based on the temporal difference between a cell's actual activation and its learned expectation, and a structural plasticity mechanism where cells physically migrate across the grid to optimize their information-receptive fields. This dual approach allows the network to learn both how to process information (synaptic weights) and also where to position its computational resources (network topology). We validated the SAPIN model on the classic Cart Pole reinforcement learning benchmark. Our results demonstrate that the architecture can successfully solve the CartPole task, achieving robust performance. The network's intrinsic drive to minimize prediction error and maintain homeostasis was sufficient to discover a stable balancing policy. We also found that while continual learning led to instability, locking the network's parameters after achieving success resulted in a stable policy. When evaluated for 100 episodes post-locking (repeated over 100 successful agents), the locked networks maintained an average 82\% success rate.
\end{abstract}


\section{Introduction}
A central challenge in understanding intelligence is explaining how stable, adaptive behavior emerges from systems that are fundamentally self-organizing rather than explicitly programmed. Biological organisms maintain internal order by continuously predicting and counteracting deviations from expected sensory states. This research draws inspiration from two key areas of theoretical and experimental neuroscience:

\begin{enumerate}
    \item \textbf{The Free Energy Principle \& Active Inference:} This framework posits that biological agents act to minimize a form of prediction error or surprise (variational free energy) \cite{friston2010, parr2019}. Perception and learning are cast as processes of updating an internal generative model to better predict sensory inputs. Action, in turn, is the process of sampling the environment to make sensations conform to predictions.
    \item \textbf{Biological Morphogenesis and Somatic Computation:} Michael Levin posits that morphogenesis itself is a form of collective intelligence \cite{levin2019}. In this view, \textit{somatic} (non-neural) cells form bioelectric networks that store and process information, enabling them to work towards anatomical goals like organ construction and regeneration \cite{levin2021}. This suggests that structural plasticity is a fundamental, scale-invariant component of biological problem-solving. The concept of embodied, structural adaptation is further extended by Cortical Labs \cite{kagan2022}, which has demonstrated that dissociated \textit{neural} cultures can physically reorganize themselves.
\end{enumerate}

The Structurally Adaptive Predictive Inference Network (SAPIN) is proposed as a computational model to bridge this gap. We hypothesize that by endowing computational agents with the ability to physically move and reposition their cells, they can actively structure their own input streams, leading to more efficient learning.

The SAPIN architecture integrates two distinct learning mechanisms operating in parallel:
\begin{enumerate}
    \item \textbf{Synaptic Plasticity:} A local, Hebbian-like learning rule updates a cell's directional connection strengths ($s$) and its homeostatic activation expectation ($E$). This update is driven entirely by the local \textit{prediction error}, the difference between the cell's actual activation ($V$) and its learned expectation ($E$).
    \item \textbf{Structural Plasticity:} A novel movement mechanism allows processing cells to physically migrate across the 2D grid. This movement is driven by the long-term average prediction error, or desire. A cell that is chronically over- or under-activated relative to its expectation will move to a new location to find a more predictable position.
\end{enumerate}

We test this architecture on the Cart Pole balancing task \cite{barto-1983}. This environment serves as a canonical benchmark for homeostasis, the goal is to maintain a stable state (an upright pole) against perturbations. Our findings demonstrate that SAPIN can solve this task. Additionally, we show that the network learns to maintain homeostasis without any external reward or punishment signal. The intrinsic drive to minimize local prediction errors is sufficient, providing a computational proof-of-concept for active inference as a self-sufficient driver for this class of control problems.

\section{Theoretical Foundations and Related Work}

\subsection{The Brain as a Generative Model: Predictive Inference as a Unifying Principle}
The pursuit of designing more advanced artificial general intelligence has increasingly led researchers to re-examine the foundational principles of computation in the one system known to possess it: the biological brain. A powerful and unifying perspective emerging from theoretical neuroscience is that the brain is an active, generative inference engine rather than a passive information processor. This view posits that the brain's primary function is to construct and maintain an internal model of the world, which it uses to predict the causes of its sensory inputs.

At the highest level of abstraction is the Free Energy Principle (FEP), a comprehensive theory of brain function and, more broadly, of any self-organizing system. Proposed by Karl Friston \cite{friston2010}, the FEP posits that for any biological agent to maintain its structural and functional integrity in a constantly changing and entropic world, it must minimize a quantity known as variational free energy. Variational free energy is an information-theoretic quantity that serves as a tractable upper bound on surprise. Therefore, by minimizing free energy, an agent implicitly maximizes the evidence for its own existence, a process Friston terms self-evidencing \cite{butz2017}.

This single imperative, to minimize free energy, provides a remarkably unified account of brain function, casting perception, learning, and action as different facets of the same underlying process: approximate Bayesian inference \cite{smith2022}. Perception becomes the process of updating the agent's beliefs about the hidden causes of its sensations to minimize the discrepancy between what is predicted and what is sensed (i.e., prediction error). Learning is the process of updating the parameters of the generative model itself to provide better long-term predictions.

The enactive corollary of the FEP is Active Inference, which extends this principle to action \cite{parr2019}. Under active inference, actions are selected to minimize expected future free energy rather than to maximize an external reward signal. This means an agent will actively sample its environment to make sensations conform to its predictions, thereby minimizing future surprise. This formulation elegantly unifies two fundamental drivers of behavior: the drive to seek out preferred, goal-related states and the drive to reduce uncertainty about the world (e.g., epistemic foraging or exploration).

If the FEP and Active Inference describe the objective function that the brain optimizes, then Predictive Coding (PC) is the leading candidate for the algorithm the brain uses to perform this optimization \cite{rao1999, millidge2022}. The core idea of PC is that the brain's cortical hierarchy is engaged in a continuous, bidirectional exchange of signals to minimize prediction error \cite{rao1999}. Higher levels of the hierarchy generate top-down predictions about the activity in lower levels. These predictions are then compared with the actual bottom-up signals. The discrepancy between the prediction and the signal constitutes a prediction error. This error is the only information that is propagated up the hierarchy, where it is used to update the beliefs (or neural representations) at the higher level to produce a better prediction. This recursive process continues until prediction error is minimized, at which point the system has settled on the most likely explanation for its sensory input. This message-passing scheme maps with remarkable fidelity onto the known anatomy and physiology of the neocortex, often referred to as the canonical microcircuit \cite{isomura2022}. Mathematically, the dynamics of a PC network can be shown to be performing gradient descent on the variational free energy \cite{whittington2017}.

\subsection{Learning in Silico: The Quest for Biologically Plausible Credit Assignment}
While the predictive inference framework provides a compelling account of the objective of brain computation, a critical question remains: how does the brain solve the credit assignment problem? The dominant algorithm in modern deep learning, backpropagation, provides a powerful solution but is widely considered biologically implausible \cite{lillicrap2016, zador2019}.

The primary objections center on its violation of the constraints of local computation in neural circuits. The weight transport problem requires that the error signal be propagated backward via a feedback pathway whose synaptic weights are precisely the transpose of the feedforward weights \cite{lillicrap2016, akrout2019}. There is no known biological mechanism that could ensure such perfect symmetry. Furthermore, the update rule for a synapse in a deep layer requires access to an error signal that is computed at the output layer, violating the principle that synaptic plasticity should depend only on locally available information.

The foundational constraint for any biologically realistic learning rule is the Local Learning Principle. This principle states that the change in a synaptic weight can only depend on variables that are available locally at that synapse, both in space and time. The archetypal local learning rule is Hebbian plasticity, cells that fire together, wire together \cite{legenstein2008}. In its modern form, Spike-Timing-Dependent Plasticity (STDP), the change in synaptic strength depends on the precise relative timing of pre- and post-synaptic spikes \cite{izhikevich2006}.

To bridge the gap to goal-directed learning, neuroscience has identified a broader class of neo-Hebbian three-factor learning rules \cite{legenstein2008}. These rules augment the two local factors (pre- and post-synaptic activity) with a third, globally broadcast signal, typically a neuromodulator like dopamine. This third factor conveys information about the outcome of behavior such as reward, punishment, novelty, or surprise and acts as a gate, modulating the Hebbian plasticity. This mechanism allows the network to solve the temporal credit assignment problem, linking past neural activity to delayed behavioral outcomes.

Several algorithms, such as Feedback Alignment (FA) \cite{lillicrap2016}, Reciprocal Feedback (RF), and Equilibrium Propagation (EP) \cite{scellier2017}, have been proposed to approximate the gradient-based learning of backpropagation using only local rules. These approaches, along with Predictive Coding \cite{millidge2022}, demonstrate that effective credit assignment is possible without violating biological constraints.

\subsection{The Dynamic Blueprint: Structural Plasticity and Topographic Self-Organization}
The computational principles of predictive inference and local learning address the functional aspects of brain computation. However, they operate on a physical substrate that is far from static. The Spatially Embedded nature of the SAPIN architecture is motivated by the understanding that the brain's physical structure is a crucial and dynamic component of its computational power.

The traditional view of the adult brain as a fixed, hard-wired machine has been overturned. The brain exhibits a remarkable capacity for structural reorganization throughout its entire lifespan \cite{butz2020}. This lifelong structural plasticity manifests in several forms, including dendritic and axonal remodeling, synaptogenesis and pruning, and even adult neurogenesis in specific regions. These structural changes are influenced by an organism's experiences. This suggests that the brain's architecture is itself a form of memory, a dynamic blueprint that is continuously optimized to meet the computational demands of the environment.

Inspired by these findings, computational models of structural plasticity have been developed \cite{butz2020, galluppi2015}. Activity-dependent rules link the formation and elimination of synapses to the neural activity they support, often via a use it or lose it principle. Homeostatic rules aim to maintain a stable level of overall activity, driving a neuron to create or retract connections to restore its firing rate to a homeostatic set-point \cite{butz2020}. Incorporating these dynamic rewiring mechanisms has been shown to increase learning speed and memory storage capacity.

Beyond dynamic rewiring, another key structural feature of the brain is its spatial organization into topographic maps, where the physical arrangement of neurons reflects the structure of the sensory input.

The classic computational model for this is the Self-Organizing Map (SOM), or Kohonen map \cite{kohonen1kind}, an unsupervised network that uses competitive and cooperative learning to create a feature map that reflects the topology of the data space. Recent breakthroughs, such as Credit-Based Self-Organizing Maps (CB-SOMs) \cite{flesch2022}, have successfully integrated this principle with top-down, error-driven learning, demonstrating that the brain's spatial organization is itself shaped by the same error-minimizing imperatives that drive learning.

This principle of a dynamic physical blueprint extends far beyond the nervous system. The work of Michael Levin has powerfully demonstrated that morphogenesis, the process by which an organism develops its shape, is itself a form of computation executed by a collective intelligence of non-neural somatic cells \cite{levin2019}. All cells in the body communicate via bioelectric networks, using ion channels and gap junctions to form a cognitive glue that stores a memory of the organism's target morphology. This bioelectric software guides the deployment of the genomic hardware to achieve complex anatomical goals, such as constructing an eye or regenerating a limb \cite{levin2021}. This reveals that the optimization of physical structure to meet functional goals is a scale-invariant principle of biological intelligence, providing a deep motivation for exploring structural plasticity in computational models like SAPIN.

\subsection{Embodied Inference: From Theory to Action}
Intelligence is an embodied and enactive process that unfolds through an agent's interaction with its environment. Active Inference offers a fundamentally different and more integrated perspective on control than traditional Reinforcement Learning (RL) \cite{parr2019, smith2022}. An active inference agent seeks to minimize its Expected Free Energy (EFE). This EFE objective can be decomposed into two components: an instrumental value (exploitation, or seeking preferred outcomes) and an epistemic value (exploration, or reducing uncertainty). By casting action selection as inference, active inference provides a first-principles solution to the exploration-exploitation dilemma.

While computational models provide compelling proof-of-concept, a powerful validation of the Free Energy Principle as a basis for learning comes from the DishBrain experiment by Cortical Labs \cite{kagan2022}. This provided the a demonstration that a culture of living neurons in a dish could learn to perform a goal-directed task (playing Pong) in a simulated environment.

The learning mechanism did not involve any conventional reward signal. Instead, it was based entirely on the principle of surprise minimization. When the culture successfully hit the ball, it received a simple, predictable electrical stimulus. When it missed, it received a prolonged, chaotic, and unpredictable stimulus. In precise accordance with the FEP \cite{kagan2022}, the neural culture spontaneously organized its activity to make its sensory world more predictable by learning to return the ball, thereby avoiding the surprising chaotic feedback. This provides powerful, tangible evidence that the FEP is a real, physical principle that can drive learning and self-organization in biological neural tissue.

This principle of embodied, goal-directed behavior driven by local rules also exists outside of neural tissue. Further evidence comes from developmental bioelectricity, where cellular collectives solve complex anatomical problems. For example, by manipulating the endogenous bioelectric gradients in planarian flatworms, the memory of the organism's target morphology can be permanently rewritten, inducing a fragment to regenerate into a stable, two-headed worm \cite{levin2021}. Similarly, the creation of Xenobots, novel, living machines built from frog skin and heart cells, demonstrates that somatic cells, when removed from their normal context, can self-organize to perform new functions like locomotion and collective object manipulation \cite{kriegman2020}. These examples underscore that the fundamental principles of embodied agency and collective problem-solving are deeply conserved across diverse biological substrates, justifying their exploration in non-traditional computational architectures.

SAPIN is situated at the confluence of these research streams, synthesizing predictive inference, local learning, and dynamic structural plasticity into a single, unified architecture.

\section{Model Architecture}

\subsection{System Overview}
The SAPIN model is instantiated on a 2D grid of dimensions $9 \times 9$. The system consists of three distinct cell populations:
\begin{itemize}
    \item \textbf{Input Cells ($\mathcal{I}$):} A fixed set of 4 cells.
    \item \textbf{Processing Cells ($\mathcal{P}$):} A dynamic population of 30 cells.
    \item \textbf{Output Cells ($\mathcal{O}$):} A fixed set of 2 cells.
\end{itemize}
Most numerical values used throughout the system including inputs, activations and weights are constrained to the range $[-1, 1]$.

\subsection{Spatial Configuration}
The placement of input and output cells is fixed, while processing cells are dynamic.
\begin{itemize}
    \item \textbf{Input Locations:} The 4 input cells are fixed in the leftmost column at coordinates: (0,1), (0,3), (0,5), and (0,7).
    \item \textbf{Output Locations:} The 2 output cells are fixed in the rightmost column at coordinates: (8,2) and (8,6).
    \item \textbf{Processing Locations:} The 30 processing cells are assigned random, unoccupied (x,y) coordinates upon initialization.
\end{itemize}

\begin{figure}[t]
    \centering
    \includegraphics[width=0.9\columnwidth]{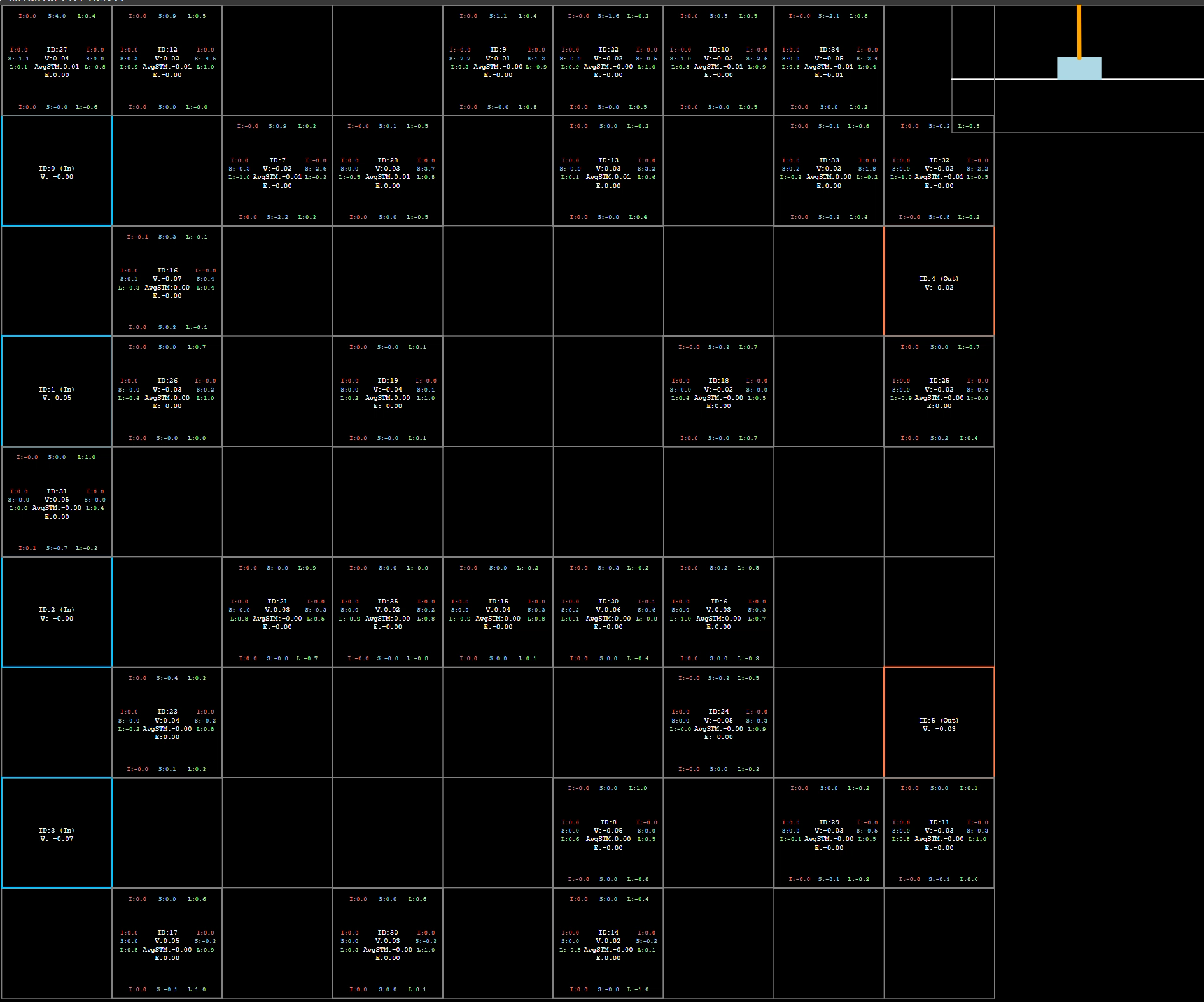}
    \caption{\textbf{Grid Setup.} The SAPIN 9x9 grid architecture. The 4 Input Cells (blue) are fixed, as are the 2 Output Cells (orange). The 30 Processing Cells (grey) are randomly initialized and migrate via structural plasticity. The top-right corner depicts the current state of the cart pole environment}
    \label{fig:grid}
\end{figure}

\subsection{Cell State Representation}
Each processing cell $i \in \mathcal{P}$ maintains three sets of variables. Input and output cells only maintain an immediate state.

\begin{itemize}
    \item \textbf{Long-Term Memory (LTM):} These variables represent the cell's learned knowledge and are updated by synaptic plasticity.
    \begin{itemize}
        \item \textbf{Directional Strengths ($s_i$):} A 4-element vector $s_i = [s_{\text{up}}, s_{\text{right}}, s_{\text{down}}, s_{\text{left}}]$ representing the cell's learned connectivity bias in each cardinal direction. Initialized with random values in $[-1, 1]$.
        \item \textbf{Expectation ($E_i$):} A scalar value representing the cell's homeostatic baseline, or its predicted activation level. Initialized with a random value in $[-1, 1]$.
    \end{itemize}
    \item \textbf{Short-Term Memory (STM):} These variables accumulate information over a macro-episode (defined as 4 full environment episodes) and are used to drive structural plasticity.
    \begin{itemize}
        \item \textbf{Accumulated Directional Influx ($v_i^{\text{saved}}$):} A 4-element vector that accumulates the directional inputs $v_i$ from each propagation wave.
        \item \textbf{Accumulated Total Influx ($V_i^{\text{saved}}$):} A scalar that accumulates the total activation $V_i$ from each propagation wave.
    \end{itemize}
    \item \textbf{Immediate State:} These variables represent the cell's activation during a single propagation wave and are reset at every environment timestep.
    \begin{itemize}
        \item \textbf{Directional Value ($v_i$):} A 4-element vector accumulating influence from the four cardinal directions during the current wave.
        \item \textbf{Total Value ($V_i$):} The cell's total activation for the current timestep.
    \end{itemize}
\end{itemize}

Output cells have a fixed, uniform directional strength $s_i = [0.25, 0.25, 0.25, 0.25]$.

\section{Network Dynamics and Adaptation}

\subsection{Signal Propagation and Action Selection}
At each environment timestep, a wave of activation propagates through the network to determine an action. This process is detailed in Algorithm \ref{alg:propagation}.

The propagation mechanism was designed to support a spatially flexible architecture. Unlike a spiking neural network, which uses a fixed activation threshold, or a standard ANN, which uses fixed directional connections, SAPIN requires a more dynamic approach. Because cells must be free to move (structural plasticity) and the learning rule is based on comparing total activation $V_i$ to a learned expectation $E_i$, a fixed "spiking threshold" is not suitable. Therefore, we use a "winner-takes-all" propagation order: the un-activated cell with the highest absolute activation $|V_j|$ is the next to propagate its signal. This allows information to flow through the grid in a data-driven, non-serial manner without pre-defined layers or static connections.

\begin{figure}[t]
    \centering
    \includegraphics[width=0.9\columnwidth]{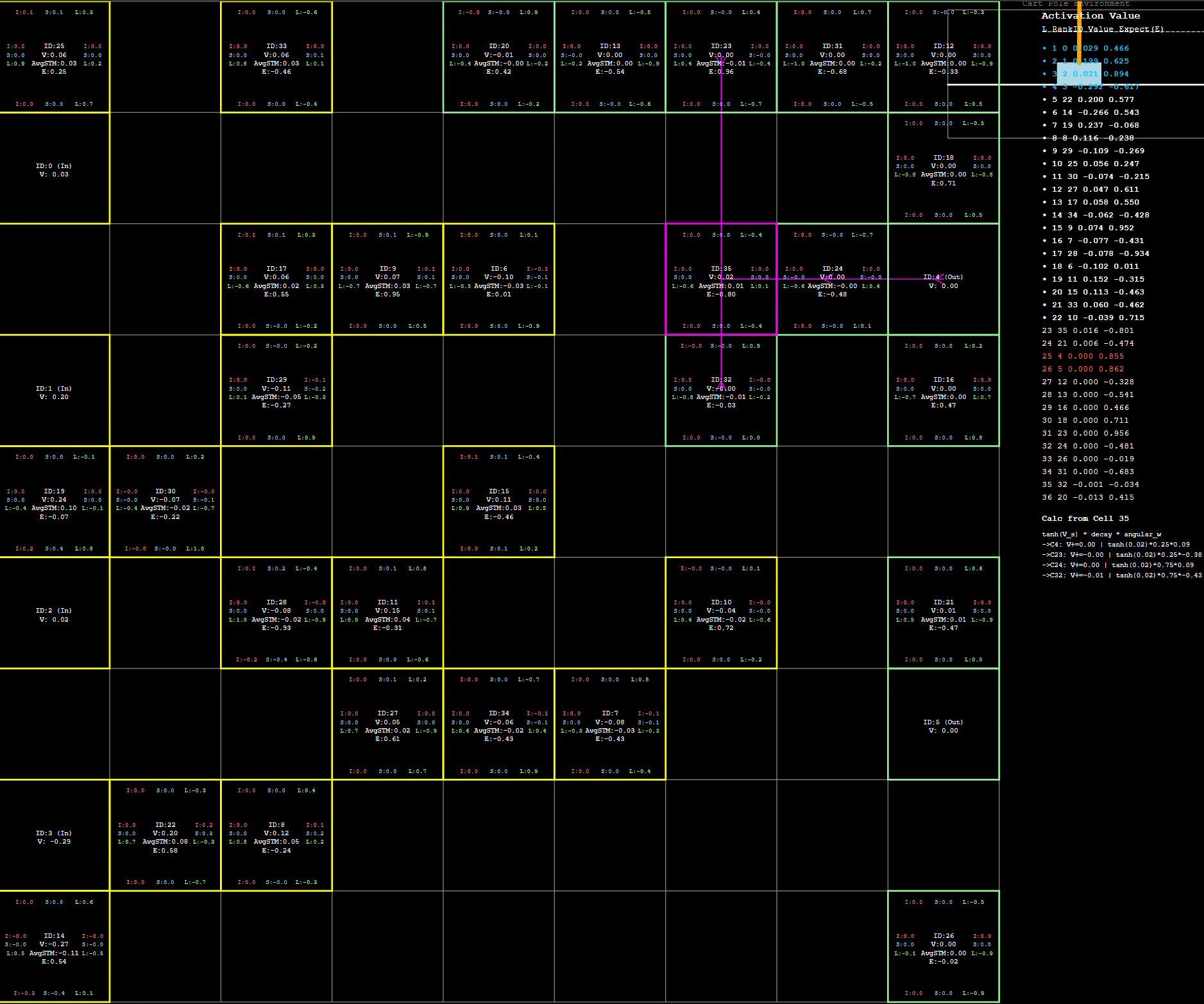}
    \caption{\textbf{Signal Propagation (Update).} Information propagates from the input cells (left) to the output cells (right). The next cell to propagate is the one with the highest absolute activation $|V|$. Nonlinearity is introduced via $\tanh(V)$ and the trigonometric angular weighting. In this visual, the cell that is currently propagating information is shown in pink. There are pink arrows pointing to the nearby cells which it is sending information to. The cells that have already been activated are colored yellow. The cells that have not yet been activated are colored green. The right column depicts the current state of the cart pole environment above a list of the value of each cell, with the value of the input cells in blue and that of the output cells in red. Below this can be viewed the expression for the current equation to propagate value from one cell to the nearby cells.}
    \label{fig:updates}
\end{figure}

\begin{algorithm}[tb] 
\caption{Signal Propagation Wave}
\label{alg:propagation}
\begin{algorithmic}[1]
    \State \textbf{Input:} Normalized state vector $S = [s_0, s_1, s_2, s_3]$
    \State \textbf{Initialize:}
    \State Reset $v_i, V_i \leftarrow 0$ for all $i \in \mathcal{P} \cup \mathcal{O}$
    \State $\mathcal{A} \leftarrow \emptyset$ (Active Set)
    \State $\mathcal{U} \leftarrow \mathcal{P} \cup \mathcal{O}$ (Available Set)
    \For{$j \leftarrow 0$ to 3} \Comment{Seed Input Cells}
        \State Let $i$ be the $j$-th input cell in $\mathcal{I}$
        \State $V_i \leftarrow S_j$
        \State Add $i$ to $\mathcal{A}$
    \EndFor
    
    \While{$\mathcal{U}$ is not empty}
        \State $\mathcal{C} \leftarrow \emptyset$ (Contribution Map)
        \For{each sender $s \in \mathcal{A}$}
            \If{$|V_s| = 0$} \textbf{continue} \EndIf
            \For{each receiver $r \in \mathcal{U}$}
                \State $d \leftarrow \text{ManhattanDistance}(s, r)$
                \State $D_d \leftarrow \text{GetDistanceDecay}(d)$
                \If{$D_d = 0$} \textbf{continue} \EndIf
                \State $W_{\text{ang}}, W_{\text{dir}} \leftarrow \text{GetAngularWeighting}(s, r)$
                \State $\Delta V \leftarrow \tanh(V_s) \cdot D_d \cdot W_{\text{ang}}$
                \State $\mathcal{C}[r] \leftarrow \mathcal{C}[r] + (\Delta V, \Delta V \cdot W_{\text{dir}})$
            \EndFor
        \EndFor
        
        \For{each receiver $r$, ($\Delta V_{\text{total}}$, $\Delta v_{\text{total}}$) in $\mathcal{C}$}
            \State $V_r \leftarrow V_r + \Delta V_{\text{total}}$
            \State $v_r \leftarrow v_r + \Delta v_{\text{total}}$
        \EndFor
        
        \If{$\mathcal{U}$ is empty} \textbf{break} \EndIf
        
        \State $i_{\text{next}} \leftarrow \arg\max_{j \in \mathcal{U}} |V_j|$
        \State Move $i_{\text{next}}$ from $\mathcal{U}$ to $\mathcal{A}$
    \EndWhile
    
    \State \textbf{Action Selection:}
    \State Let $o_0, o_1$ be the two output cells in $\mathcal{O}$
    \State \textbf{return} $0$ if $V_{o_0} > V_{o_1}$ else $1$
\end{algorithmic}
\end{algorithm}

The distance decay function $D(d)$ is a discrete lookup based on Manhattan distance $d$:
$$ D(d) = \begin{cases} 
    1.0 & \text{if } d = 0 \\
    0.75 & \text{if } d = 1 \\
    0.25 & \text{if } d = 2 \\
    0.0 & \text{if } d \ge 3 
\end{cases} $$
The angular weighting $W_{\text{ang}}$ interpolates the sender's directional strengths $s_s$. For an angle $\theta$ to the receiver, the directional components (up, right, down, left) are weighted by $\max(0, -\sin\theta)$, $\max(0, \cos\theta)$, $\max(0, \sin\theta)$, and $\max(0, -\cos\theta)$, respectively. $W_{\text{ang}}$ is the dot product of $s_s$ and these calculated weights.

After the wave terminates, and if the network's global lock is not engaged, the STM variables for all processing cells are updated:
\begin{align}
    v_i^{\text{saved}} &\leftarrow v_i^{\text{saved}} + v_i \\
    V_i^{\text{saved}} &\leftarrow V_i^{\text{saved}} + V_i
\end{align}

\subsection{Synaptic Plasticity (LTM Update)}
After each action, the network updates its LTM (synaptic weights) based on the immediate activations. This process, detailed in Algorithm \ref{alg:ltm}, is skipped if the network is globally locked. The learning rate was set to $\eta = 0.02$.

\begin{algorithm}[tb] 
\caption{Synaptic Plasticity (LTM Update)}
\label{alg:ltm}
\begin{algorithmic}[1]
    \State \textbf{Input:} Learning rate $\eta = 0.02$
    \If{Network is globally locked} \textbf{return} \EndIf
    
    \For{each cell $i \in \mathcal{P}$}
        \State $\text{error}_i \leftarrow V_i - E_i$
        
        \State \Comment{Update Expectation}
        \State $E_i \leftarrow E_i + (\eta / 2) \cdot \text{error}_i$
        
        \State \Comment{Update Directional Strengths}
        \State $v_{\text{sum}} \leftarrow \sum_{d \in \text{dirs}} |v_{i,d}|$
        \If{$v_{\text{sum}} > 10^{-6}$}
            \State $p_i \leftarrow v_i / v_{\text{sum}}$ \Comment{Directional Proportions}
            \State $s_i \leftarrow s_i + (\eta / 2) \cdot p_i \cdot \text{error}_i$
            \State $s_i \leftarrow \text{clip}(s_i, -1, 1)$
        \EndIf
    \EndFor
\end{algorithmic}
\end{algorithm}

\subsection{Structural Plasticity (Movement)}
Structural plasticity occurs after a macro-episode, which is defined as 4 full environment episodes. This process (Algorithm \ref{alg:movement}) allows cells to physically relocate. It is also skipped if the network is globally locked.

The movement logic is driven by the same error-minimizing principle as synaptic plasticity. A cell's "desire" to move is its long-term average prediction error. The direction of movement is determined by the source of the unexpected input. A cell moves along the axis from which it received the most "surprising" (highest magnitude) average influx, as captured by the weighted choice based on $\bar{v}_i$. If over-activated, it moves away from that source; if under-activated, it moves toward it. Inspired by reinforcement learning, a small random chance ($\epsilon_{\text{rand}}$) is included to encourage exploration and prevent the network from getting "stuck" in a poor, but stable, spatial configuration.

\begin{figure}[t]
    \centering
    \includegraphics[width=0.9\columnwidth]{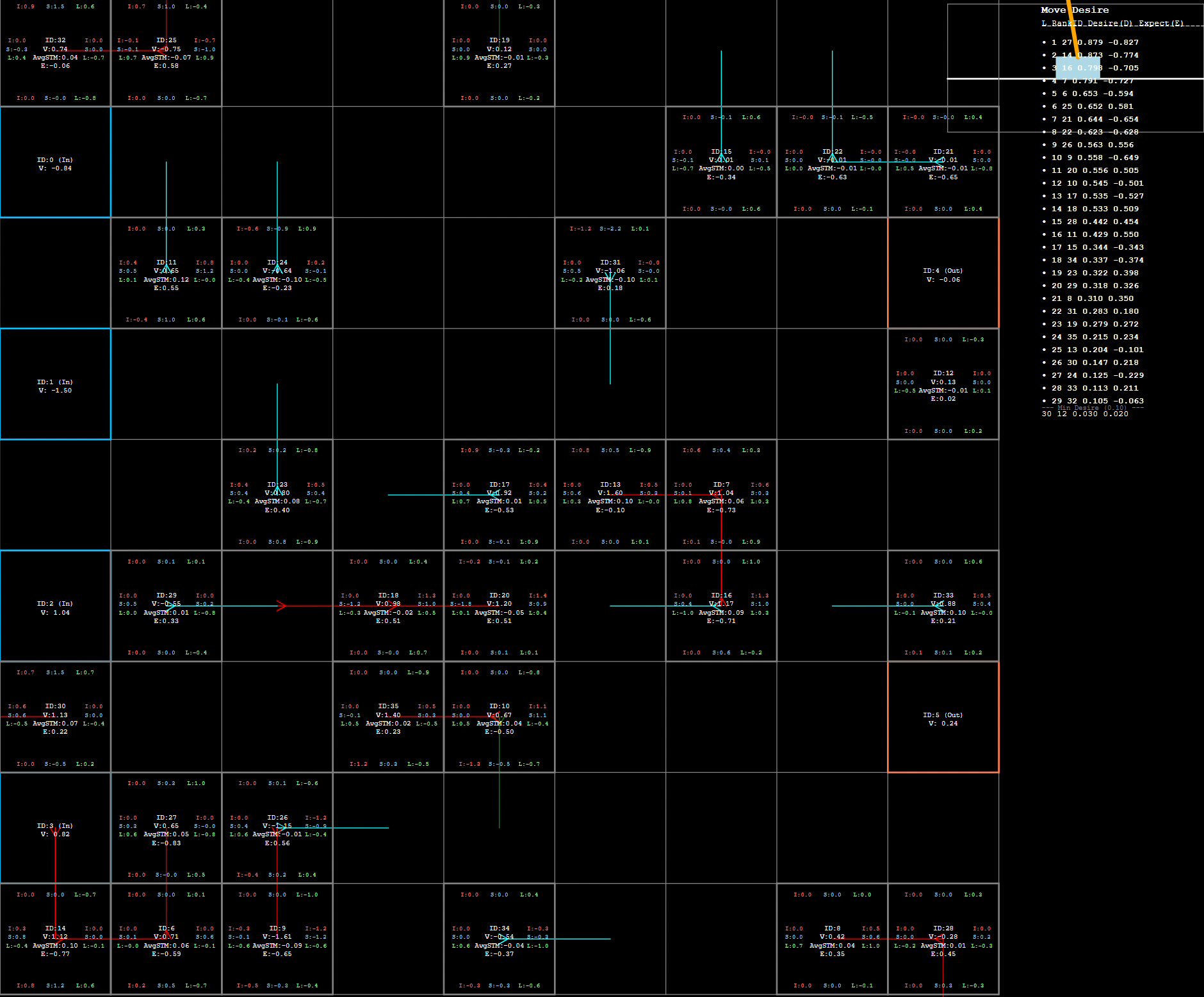}
    \caption{\textbf{Structural Plasticity (Movement).} Cells with high long-term prediction error ($|\bar{V}_i - E_i|$) move to a new location. If over-activated, a cell moves away from its main signal source; if under-activated, it moves toward it. The blue arrows depict where a cell moves to. The red arrows depict where a cell was unable to move due to a collision. Order of movement is determined by highest desire. The right column depicts the current state of the cart pole environment above a list of the desire of each cell to move. A horizontal line cuts off the list of cells between those with a high enough desire to move and those with too low a desire to move.}
    \label{fig:movement}
\end{figure}

\begin{algorithm}[tb!] 
\caption{Structural Plasticity (Movement)}
\label{alg:movement}
\begin{algorithmic}[1]
    \State \textbf{Input:} Min desire $\theta_D=0.1$, $\epsilon_{\text{rand}}=0.05$
    \State \textbf{Initialize:} $N_{\text{steps}} \leftarrow$ steps in macro-episode
    \If{Network is globally locked \textbf{or} $N_{\text{steps}} = 0$}
        \State Reset all STM ($v^{\text{saved}}, V^{\text{saved}} \leftarrow 0$)
        \State \textbf{return}
    \EndIf
    
    \State \textbf{Calculate Desire:}
    \State $\mathcal{M} \leftarrow \emptyset$ (Movement Candidates)
    \For{each cell $i \in \mathcal{P}$}
        \State $\bar{V}_i \leftarrow V_i^{\text{saved}} / N_{\text{steps}}$
        \State $\bar{v}_i \leftarrow v_i^{\text{saved}} / N_{\text{steps}}$
        \State $D_i \leftarrow |\bar{V}_i - E_i|$
        \If{$D_i \ge \theta_D$ \textbf{or} $\text{random}() < 0.025$}
            \State Add $(i, D_i, \bar{V}_i, \bar{v}_i)$ to $\mathcal{M}$
        \EndIf
    \EndFor
    
    \State Sort $\mathcal{M}$ by $D_i$ descending
    \State $\text{Occupied} \leftarrow$ set of all cell coordinates
    
    \For{each $(i, D_i, \bar{V}_i, \bar{v}_i)$ in $\mathcal{M}$}
        \State \Comment{1. Choose movement axis}
        \If{$\text{random}() < \epsilon_{\text{rand}}$}
            \State $\text{dir} \leftarrow \text{random\_direction}()$
        \Else
            \State $P \leftarrow |\bar{v}_i| / \sum |\bar{v}_i|$
            \State $\text{dir} \leftarrow \text{weighted\_choice}(\text{dirs}, P)$
        \EndIf
        
        \State \Comment{2. Choose movement vector (towards/away)}
        \If{$\bar{V}_i > E_i$} \Comment{Over-activated}
            \State $\text{move\_vec} \leftarrow -1 \cdot \text{get\_vector}(\text{dir})$
        \Else \Comment{Under-activated}
            \State $\text{move\_vec} \leftarrow +1 \cdot \text{get\_vector}(\text{dir})$
        \EndIf
        
        \State \Comment{3. Attempt Move}
        \State $(x_{\text{new}}, y_{\text{new}}) \leftarrow (x_i, y_i) + \text{move\_vec}$
        \If{$(x_{\text{new}}, y_{\text{new}})$ is valid \textbf{and} not in $\text{Occupied}$}
            \State Remove $(x_i, y_i)$ from $\text{Occupied}$
            \State Add $(x_{\text{new}}, y_{\text{new}})$ to $\text{Occupied}$
            \State $(x_i, y_i) \leftarrow (x_{\text{new}}, y_{\text{new}})$
        \EndIf
    \EndFor
    
    \State Reset all STM ($v^{\text{saved}}, V^{\text{saved}} \leftarrow 0$)
\end{algorithmic}
\end{algorithm}

\section{Experimental Protocol}

\subsection{Environment and Task}
The SAPIN architecture was evaluated on the CartPole-v1 environment from the Gymnasium library \cite{gymnasium-2024}. The 4-dimensional state vector (cart position, cart velocity, pole angle, pole angular velocity) was normalized to the range $[-1, 1]$ using the environment's standard bounds (cart position $\pm 2.4$, cart velocity $\pm 4.0$, pole angle $\pm 0.209$ rad, pole angular velocity $\pm 4.0$). This normalized 4D vector was mapped directly to the 4 input cells. An episode was considered successful if the agent balanced the pole for 500 consecutive timesteps.

\subsection{Punishment Mechanism}
\label{sec:punishment}
Drawing inspiration from the DishBrain's experiment's use of unpredictable stimuli \cite{kagan2022}, we implemented a punishment mechanism designed to inject surprise into the network. This was implemented in two ways:

\begin{enumerate}
    \item \textbf{Catastrophic Failure:} Upon episode termination (pole falls), 10 epicenters were created at random (x,y) grid locations. Each epicenter emitted a random value $p \in [-1, 1]$.
    \item \textbf{Probabilistic Punishment:} During non-terminal states, if the pole angle was high (between 4 and 12 degrees), there was a 1-10\% chance (scaling with the angle) of a punishment event of 1-30\% of the 10 epicenters.
\end{enumerate}

In both cases, the punishment values initiated a special propagation wave (Algorithm \ref{alg:propagation}), and the resulting cell activations $V_i$ were used to drive a synaptic update (Algorithm \ref{alg:ltm}). This was intended to update the network's LTM to avoid the states that led to the punishment.

\begin{figure}[t]
    \centering
    \includegraphics[width=0.9\columnwidth]{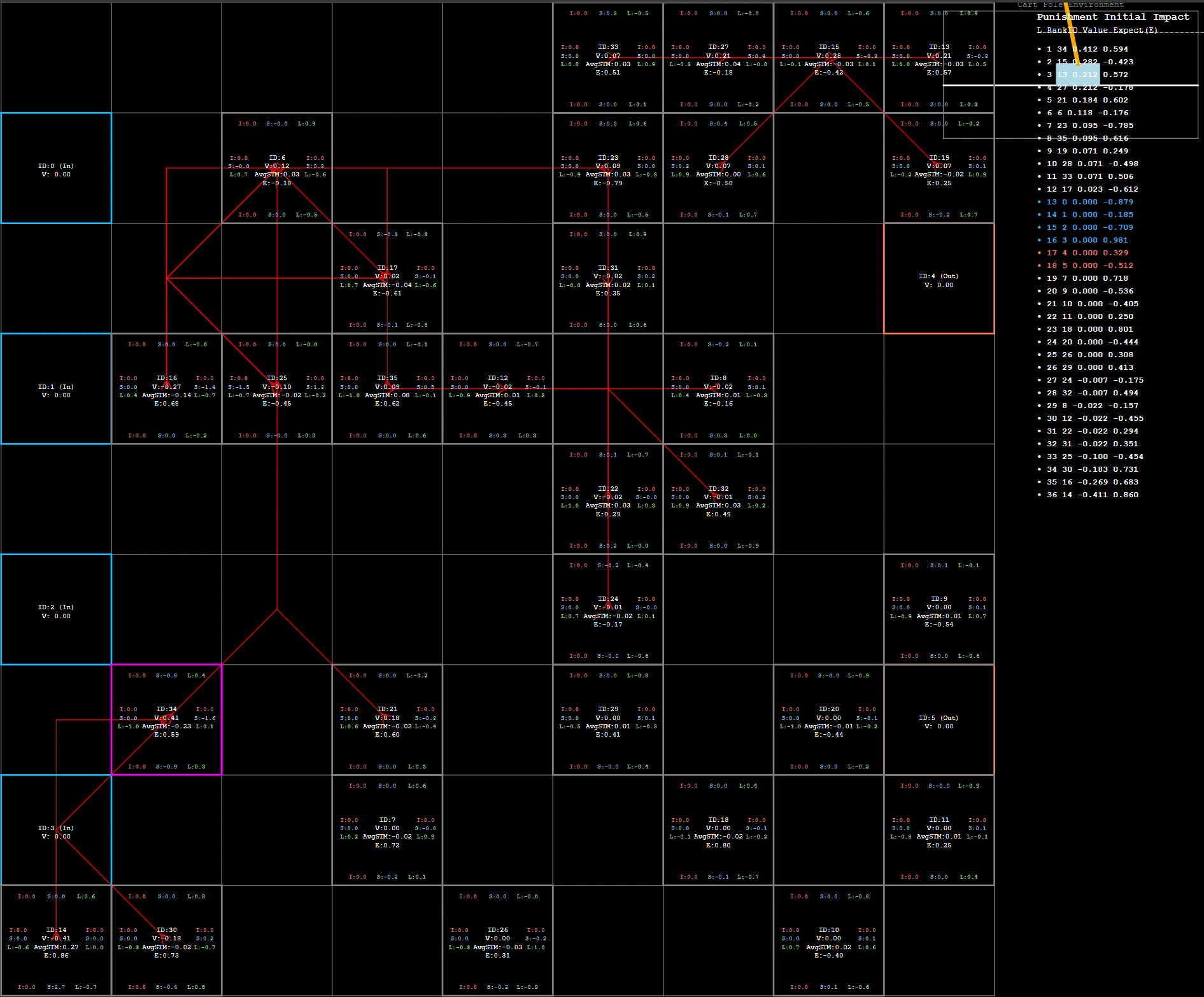}
    \caption{\textbf{The Punishment Mechanism.} Upon failure, random epicenters (at empty or non-empty cells) generate chaotic signals that propagate through the network, driving a synaptic update intended to associate the preceding states with surprise. In this visual, red arrows are shown from each of the random epicenters and connect to the nearby cells. Then a propagation wave happens like normal. The next cell to propagate is the one with the farthest value form 0, like normal. The first cell is shown in pink.}
    \label{fig:punishment}
\end{figure}

\section{Results and Analysis}

\subsection{Performance and Instability}
The SAPIN network proved highly successful at solving the Cart Pole environment. It frequently achieved success (500 steps) within the first 10 episodes. The agent's behavior demonstrated clear corrective actions to balance the pole.

However, this success was often unstable. A network that achieved 500 steps might, in the very next episode, fail after only 10 steps. This is attributed to the fact that the learning rules (Algorithms \ref{alg:ltm} and \ref{alg:movement}) do not guarantee convergence. The network is constantly adapting, and a good policy could be forgotten as the network continues to explore its state space. We also observed that poor random initial positions for the processing cells could prevent the agent from succeeding for 100 episodes or more.

\subsection{The Locking Experiment}
To address the instability, we experimented with locking the network's parameters. A global boolean flag was added to the SAPIN model. When the agent first achieved a score of 500 steps, this flag was set to True, permanently disabling all LTM updates (Algorithm \ref{alg:ltm}) and all structural plasticity (Algorithm \ref{alg:movement}).

To test the stability of a found policy, the network was evaluated for 100 episodes immediately after locking. This process was repeated for 100 different successfully trained agents. On average, the locked networks maintained an \textbf{82\% success rate} (i.e., 82 out of 100 episodes lasted the full 500 steps). This demonstrates that the SAPIN architecture is fully capable of finding and storing a robust, generalizable policy for the Cart Pole task. This 'locking' mechanism serves as a computational analogue to synaptic consolidation, a process where new memories are stabilized by a reduction in plasticity. This finding suggests a potential mechanism for resolving the well-known stability-plasticity dilemma.

\subsection{The Surprising Role of Punishment}
A finding of this research relates to the punishment mechanism described in Section \ref{sec:punishment}. We ran a comparison of three experimental conditions:
\begin{enumerate}
    \item Punishment on catastrophic failure only.
    \item Punishment on catastrophic failure and probabilistic punishment during poor performance.
    \item No punishment whatsoever.
\end{enumerate}

Counter to our initial hypothesis, all three conditions produced very similar results. The network successfully learned to balance the pole even when it was never punished for failing.

This strongly implies that, for a homeostatic task like Cart Pole, the agent's intrinsic drive to minimize its own local prediction errors is sufficient for learning. The network learns to maintain homeostasis (keep the pole balanced) because a balanced pole provides a stable, and therefore highly predictable, stream of sensory input. Dropping the pole, by contrast, results in a chaotic and unpredictable sensory state. The network learns to seek the state of minimal prediction error, which in this environment, is the success state.

This finding suggests that the core mechanism of active inference, minimizing surprise, can be a sufficient objective function for certain control tasks, without any need for an externally defined reward or punishment signal \cite{friston2010}.

\subsection{Discussion and Future Work}
The homeostasis-seeking nature of the agent is both a strength and a weakness. It is perfectly suited for the Cart Pole task, which is itself a problem of maintaining homeostasis. However, this raises questions about the model's ability to solve tasks that require long-term planning or deliberately moving away from a stable state to achieve a more distant goal. This reliance on immediate prediction error is a potential limitation, consistent with models that only minimize immediate prediction error. Future work could address this by implementing deep active inference, where the agent learns a temporal model to minimize expected future free energy, enabling it to sacrifice short-term homeostasis for long-term goals.

Our attempts to use a punishment signal to guide the agent had no noticeable effect. Future work should investigate why this signal was ineffective. It may be that the local, homeostatic updates (Algorithm \ref{alg:ltm}) are stronger than the updates from the punishment wave, or that the random nature of the punishment signal was too noisy to provide a useful learning gradient. We also tested an alternative structural plasticity rule where, instead of moving to reduce the error magnitude, cells moved to a location with the *smallest variation* in error. The goal was to seek predictability, even if the mean activation did not match the expectation. This approach was less successful than the default rule, suggesting that matching a homeostatic set-point is a more effective drive.

An alternative to improving learning from bad initial states would be to implement a genetic algorithm to evolve the optimal initial positions of the processing cells, which would then be fine-tuned by the plasticity mechanisms.

Future work will also include evaluating this network on more complex tasks that require multi-step planning to evaluate whether homeostasis can be maintained long-term over changing environments.

The state for each cell is currently represented by 5 values: a single expectation and four directional strengths. Future work could focus on giving the cell a larger look-up table with bins connecting inputs to specific values.

Future work will extend this by implementing a continuous rather than discrete system for SAPIN. This will decrease the step size when the cells move, greatly stabalizing the model.

Additionally, future work will explore more complex structures, such as configurations wrapped around a cylinder rather than on a flat grid. 

One of the interesting aspects of biological systems is that they are deformed by their own computation. We will integrate this changing structure into the environment more closely. In the example of the cart pole environment, this will be achieved by placing the cells onto the pole rather than being separate. Thus, as the cells move, their location will directly impact the physics of the environment.

\section{Conclusion}
We introduced SAPIN, a novel, biologically-inspired computational architecture that synthesizes two forms of plasticity: local, error-driven synaptic plasticity and global, desire-driven structural plasticity. The model is grounded in the principles of Active Inference and the Free Energy Principle.

We successfully demonstrated that this architecture can solve the Cart Pole benchmark. The intrinsic objective to minimize local prediction error (i.e., to seek homeostasis) was a sufficient driver for discovering the correct policy. We also showed that while the network's continual plasticity creates instability, a locked version of a successful network provides a stable and highly robust policy.

SAPIN serves as a computational proof-of-concept for models that learn not only how to process information, but where to position their computational resources, grounding abstract inference in a dynamic, physical substrate.

\bibliography{main}

\end{document}